\documentclass[letterpaper, 10 pt, conference]{ieeeconf}

\IEEEoverridecommandlockouts 

\usepackage[utf8]{inputenc}
\usepackage[euler]{textgreek}
\usepackage{graphicx}
\graphicspath{ {figures/} }
\usepackage{graphicx}
\usepackage{float}
\usepackage{tikz}
\usepackage{amsmath,hyperref}
\usepackage{amssymb}
\usepackage{algpseudocode}
\usepackage{algorithm}
\usepackage{epstopdf}

\usepackage{multirow}

\usepackage{subfig}
\usepackage{balance}
\usepackage{color}

\usepackage{bbm}

\begin{document}
%
\title{ROMANO: A Novel Overlay Lightweight Communication Protocol for Unified Control and Sensing of a Network of Robots}

\author{\authorblockN{Pradipta Ghosh, Jason A. Tran, Daniel Dsouza, Nora Ayanian, and Bhaskar Krishnamachari}
\authorblockA{University of Southern California, Los Angeles, CA 90007, USA, \\ Email: pradiptg@usc.edu, jasontra@usc.edu, dmdsouza@usc.edu, ayanian@usc.edu, bkrishna@usc.edu}}


%


\maketitle
\begin{abstract}

We present the Robotic Overlay coMmunicAtioN prOtocol (ROMANO), a lightweight, application layer overlay communication protocol for a unified sensing and control abstraction of a network of heterogeneous robots mainly consisting of low power, low-compute-capable robots. ROMANO is built to work in conjunction with the well-known MQ Telemetry Transport for Sensor Nodes (MQTT-SN) protocol, a lightweight publish-subscribe communication protocol for the Internet of Things and makes use its concept of ``topics'' to designate the addition and deletion of communication endpoints by changing the subscriptions of topics at each device. We also develop a portable implementation of ROMANO for low power IEEE 802.15.4 (Zigbee) radios and deployed it on a small testbed of commercially available, low-power, and low-compute-capable robots called Pololu 3pi robots. Based on a thorough analysis of the protocol on the real testbed, as a measure of throughput, we demonstrate that ROMANO can guarantee more than a $99.5\%$ message delivery ratio for a message generation rate up to 200 messages per second. The single hop delays in ROMANO are as low as 20ms with linear dependency on the number of robots connected. These delay numbers concur with typical delays in 802.15.4 networks and suggest that ROMANO does not introduce additional delays. Lastly, we implement four different multi-robot applications to demonstrate the scalability, adaptability, ease of integration, and reliability of ROMANO.
\end{abstract}



%
\IEEEpeerreviewmaketitle

\section{Introduction}

With the recent advances in cheap, scalable, low power, and radio equipped robots with
small form factors, the fields of swarm robotics and collaborative networks of
robots have grabbed significant attention from wireless networks and the robotics
researchers. 
Over the last decade, groups of robots have been employed (or envisioned) in a wide range of contexts including disaster relief~\cite{ollero2003helicopter}, search and rescue missions \cite{penders2011robot}, temporary wireless communication backbone deployments~\cite{williams2014route}, 
and extending existing communication infrastructures~\cite{dhekne2017extending}. 
Nonetheless, researchers have recently noticed a lack of communication protocols engineered specifically for efficient control and data collection in a network of heterogeneous robots mainly consisting of low power robots with low computational capabilities.

The most popular state-of-the-art method for effective control of robots and effective
collection of sensor data relies on the Robot Operating System (ROS)~\cite{ros}. 
However, the traditional ROS-based solutions require high enough compute power to run a full-fledged Linux OS.
Moreover, ROS uses XML-RPC which relies on HTTP, a protocol with large header sizes that in turn consume more
bandwidth and power. 
This makes ROS unsuitable and unoptimized for a network of battery-operated robots operating with low-power embedded processors on a \emph{shared wireless communication channel}.

\textbf{Our Contribution: } We propose the Robotic Overlay coMmunicAtioN prOtocol (ROMANO)
which is a novel \emph{lightweight overlay networking protocol} for sensing and control of a set of heterogeneous robots
that builds on the cutting-edge lightweight publish-subscribe Internet of Things (IoT) protocol called MQ Telemetry Transport for Sensor Nodes (MQTT-SN)~\cite{mqttsn}. 
ROMANO employs the concept of ``topics'' from the MQTT-SN communication protocol to create a overlay network of robots where each robot subscribes/publishes to a set of control and sensing related topics (e.g., gyroscope, proximity, location, speed, movement control instructions, etc.).
ROMANO can change/control the topic subscriptions of different robots to control the ROMANO communication
endpoints for different types of communication: one to one, one to many, many to many, or many to one.
We implement the ROMANO protocol on a small testbed of Pololu 3pi robots that consists of an OpenMote~\cite{Openmote} which uses a Texas Instruments CC2538 System-on-Chip for 802.15.4/6LoWPAN and an ARM mbed LPC1768 board to evaluate its performance via a set of real world experiments. 
The experimental results demonstrate that ROMANO can guarantee $\geq 99.5\%$ message delivery ratio for a
message generation rate up to $200$ Messages Per Seconds (MPS) using $32$ bytes ROMANO message size with a delay as low as $\approx20$ms which increases linearly as a function of the number of the robots in the system.
Furthermore, as proof of concepts, we implement four different applications of ROMANO in a network of
robots: (1) simplified control of the robots, (2) seamless sharing of sensor data,
(3) control of any functions of a robot such as peer-to-peer radio transmission,
and (4) communication and control between multiple networks of robots over internet.

\textbf{State-of-the-art:}
The widespread alternatives to MQTT-SN are MQTT, Constrained Application Protocol
(CoAP)~\cite{bormann2012coap}, and XML-RPC (ROS).
Amaran \emph{et al.}~\cite{amaran2015comparison} presented a comparison of these protocols,
showing that MQTT-SN and CoAP have similar performance and advantages over MQTT 
and XML-RPC (ROS). Additionally, they showed that MQTT-SN messages are slightly 
more efficient than CoAP, which motivates our choice of MQTT-SN.
While MQTT-SN has been proposed and used in the context of sensing in
static IoT sensor deployments, MQTT-SN has not yet been used for a network of robots.
Many researchers have envisioned/proposed utilizing MQTT in the context of robotic control.
Aroon~\cite{aroon2016study} demonstrated the feasibility of
remotely controlling a single robot over a cloud platform via MQTT.
Kazala \emph{et al.}~\cite{kazala2015wireless} have also presented a proof of concept implementation of using
basic functionalities of MQTT for data exchange among multiple robots.
However, to the best of our knowledge, there are no publications presenting a 
low power MQTT-SN based overlay protocol that allows nodes to easily facilitate communication endpoints
(one-to-one, one-to-many, many-to-one, many-to-many) among a robot swarm. 
SENORA, proposed in \cite{rahman2007SENORA}, includes an inter-robot communication
protocol which takes into account robot location for medium access and
peer-to-peer communication, but the authors do not detail how generic
communication would be facilitated among robots (e.g. one-to-one or one-to-many) which
our protocol addresses. 
The authors of \cite{claro2014autonomous} proposed a messaging architecture for
inter-robot communication with the target application specifically for integration
into a surveillance system.
In contrast, our work targets generic multi-robot systems along with
various application-specific real implementation details.
Sauer \emph{et al.}~\cite{sauer2016using} presented the concept of an overlay protocol built on top of the
CoAP but do not present any implementation details or performance evaluation.
The lack of details makes it impossible for us to replicate and compare with the existing work.
On the other hand, ROS messages are widely used to connect robot swarms. 
For example, Yan \emph{et al.} \cite{yan2017hibot} presented a prototype system built
on ROS messages for robot communication and described the ease of scaling their
system, although they do not include any performance evaluation.
However, while ROS messages are powerful, their packet header overhead and the computation
requirement to run ROS do not make it ideal or bandwidth-efficient for low capacity 
robots consisting of only microcontrollers using IEEE 802.15.4 radios.
Nonetheless, because ROMANO is built on MQTT-SN, it is still possible to bridge nodes running
ROMANO with ROS, but messages passed will still be in the MQTT-SN format.

\section{The Proposed ROMANO Protocol}

\subsection{Preliminaries: MQTT/MQTT-SN with ROMANO}
\label{sec:prelim}

In this section, we briefly explain the core concepts of MQTT-SN and MQTT to better understand ROMANO.
MQ telemetry Transport (MQTT)~\cite{mqttsn} is a publish-subscribe based machine
to machine application layer networking protocol for the Internet of Things (IoT).
The core idea is that a set of ``subscriber'' nodes are connected to a set of 
``publisher'' nodes via a ``broker'' and the concept of a ``topic''. When there are multiple 
subscribers to a topic, the MQTT broker dispatches copies of a published message via 
sequential unicast. 
MQTT works 
on top of the Transmission Control Protocol (TCP).
MQTT also provides its own Acknowledgements (ACKs) for additional 
reliability via three different Quality of Service 
(QoS) modes numbered in the order of increasing complexity: QoS 0, QoS 1, and QoS 2. 

MQTT for Sensor Nodes (MQTT-SN) is a variant of MQTT that is focuses on resource 
constrained devices. MQTT-SN uses the User Datagram Protocol (UDP) rather than 
TCP and has smaller message headers to reduce the overall overhead. However, 
MQTT-SN still maintains reliability through the QoS levels used in MQTT 
\footnote{For a more detailed description of MQTT-SN, interested readers 
are referred to~\href{http://mqtt.org/new/wp-content/uploads/2009/06/MQTT-SN\_spec\_v1.2.pdf}{http://mqtt.org/new/wp-content/uploads/2009/06/MQTT-SN\_spec\_v1.2.pdf}}. 
ROMANO utilizes the ``Data'' field of a MQTT-SN Publish Message (See Table~\ref{table:mqtt-format}) for sending ROMANO messages between nodes and the TopicId field (containing the topic id value or short topic name) for identifying communication endpoints.
\begin{table}[!ht]
\centering
\caption{MQTT-SN Publish Message Format}

\begin{tabular}{|c|c|c|c|c|c|}
\hline
length    & Msg Type & Flags & Topic id & MsgId & Data \\ 
(octet 0) & (1)      & (2)   & (3-4)    & (5-6) & (7:n) \\ \hline 
\end{tabular}
\label{table:mqtt-format}
\end{table}



\subsection{ROMANO Protocol Description}
\label{sec:proposed}



\textbf{Requirements:} 
At minimum, ROMANO requires each end devices/robots to run a
MQTT-SN client on a multi-threaded OS. Each of the devices needs to be connected to a
MQTT-SN broker/forwarder while the broker nodes are bridged together either directly or
over the internet. Furthermore, the broker device also runs a ROMANO server program.
Each end device (robot or sensor node) needs to follow a standard connection \emph{establishment/initialization phase} to initiate ROMANO as follows.

\begin{itemize}
    \item Set up a MQTT-SN connection to a MQTT-SN broker where the device's IPV6 address is used as the device identifier.
    
    \item Subscribe to a topic named after the last 8 characters of the device's IPv6 address which we refer to as the ROMANO ID. For example, a device with address $fe80::212:4b00:abcd:1234$ subscribes to the topic ``$abcd1234$''.\footnote{One can use the whole IPv6 address as the topic. However, ROMANO uses the last 8 characters to keep ROMANO ID small while accommodating up to $2^{32}$ devices.} The ROMANO ID topic can be used to communicate to a specific device.
    
    \item Publish the ROMANO ID on a predefined topic ``init-info'' and wait for a fixed amount of time (2 seconds in our implementation) for an acknowledgment to be published by the ROMANO server on the respective ROMANO ID topic.
    If no acknowledgement is received on time, the node retries indefinitely.
    
    \item Subscribe to the topic ``common'' which ROMANO uses for broadcast communication. 
\end{itemize}

\begin{figure}[!ht]
    \centering
    \includegraphics[width=\linewidth]{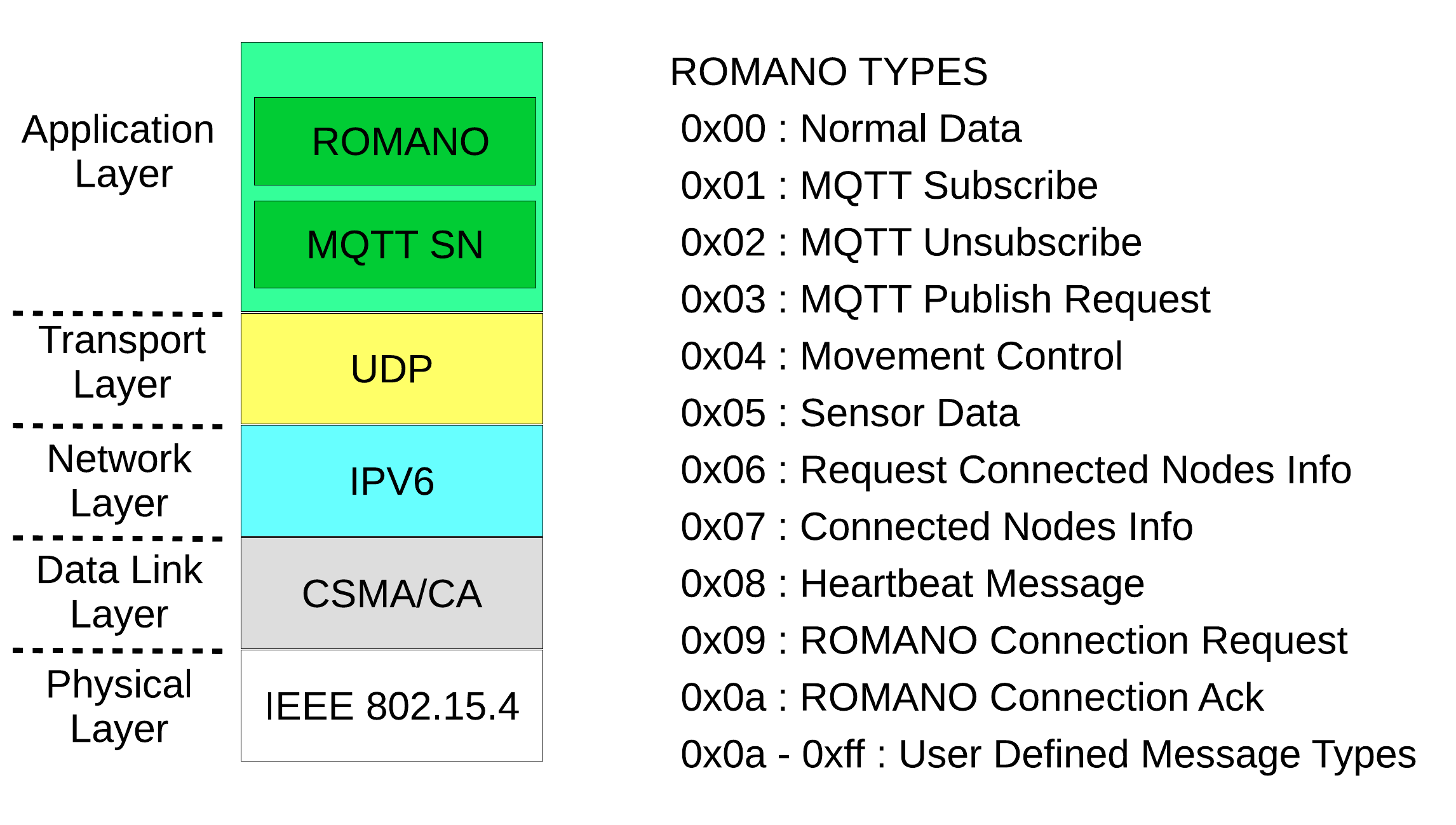}
    \caption{(Left) The ROMANO Network Stack, (Right) ROMANO Data Types}
    \label{fig:romano}
\end{figure}

\textbf{The ROMANO Protocol:}
Our proposed ROMANO protocol can be described as follows.
\begin{itemize}
    \item According to the five layered internet model of networks, the ROMANO protocol falls under the application layer alongside  the MQTT-SN protocol. More specifically, the ROMANO and the MQTT-SN protocols form a nested, layered structure inside the application layer as presented in Fig.~\ref{fig:romano}.
    
    \item ROMANO uses the MQTT-SN topics to define the communication endpoints. Any publisher to a certain topic is the transmitter node while all the subscribers of that topic are the receivers. Any node of a ROMANO network can be a transmitter at any instance of time for any topic while the receiver nodes need to subscribe first with the broker and remain connected. \emph{Thus, ROMANO allows all types of communication: one-to-one, one-to-many, many-to-one, and many-to-many.}
    
    \item ROMANO uses the MQTT-SN data section for the overlay communication where a complete ROMANO message is embedded in the MQTT-SN data section.  
    
    \item ROMANO has the feature of controlling the subscriptions of a node. One of the ROMANO message types, MQTT Subscribe can instruct the receivers to subscribe to a particular topic (say, `test-topic') so that they can register themselves as receivers of that topic (`test-topic').
    
    \item ROMANO also has the feature of instructing the receivers to publish certain types of data (e.g. telemetry data) to certain topics (e.g. `telemetry'). This feature can be used for active polling of sensor/control data for a specific robot such as the leader.
    
    \item The ROMANO overlay protocol allows any node (e.g. robot or sensor) in the network to control the movements of a robot via the same abstraction regardless of whether they are either connected directly, connected via an ad hoc network, or connected over the internet. 
    
    \item ROMANO has an optional periodic `heartbeat' messaging feature to notify its presence to all the connected nodes, which can be used for neighbor discovery or end-to-end reliable message transfer.
    
\end{itemize}

\begin{table}[!ht]
\centering
\caption{ROMANO Message Format}
\begin{tabular}{|p{1.5cm}|p{2cm}|p{3cm}|}
\hline
ROMANO Data Type  & ROMANO MSG Length  & ROMANO Data \\ 
(1 octet) & (1 octet) & (1-253 octets) \\ \hline 
\end{tabular}
\label{table:romano_msg}
\end{table}

\begin{table}[!ht]
\centering
\caption{ROMANO Message Formats}

\begin{tabular}{|p{2.5cm}|p{2.5cm}|p{2cm}|}
\multicolumn{3}{c}{ROMANO Message Format for Request Connected Nodes Info,}\\
\multicolumn{3}{c}{Heartbeat Message, and ROMANO Connection Request}\vspace{1ex} \\ \hline
ROMANO Data Type  & MSG Length & ROMANO ID \\ 
(octet 0) & (1) & (2 - 9) \\ \hline 
\end{tabular}
\vspace{2ex}

\begin{tabular}{|p{3cm}|p{3cm}|p{1cm}|}
\multicolumn{3}{c}{ROMANO Normal Data / Connected Nodes Info message Format} \vspace{1ex} \\ \hline
ROMANO Data Type  & MSG Length &  Data \\ 
(octet 0) & (1) & (2 - k) \\ \hline 
\end{tabular}
\vspace{2ex}

\begin{tabular}{|p{2cm}|p{2cm}|p{3cm}|}
\multicolumn{3}{c}{ROMANO MQTT SUB/UNSUB Control Message Format} \vspace{1ex} \\ \hline
ROMANO Data Type & MSG Length & Topic to subscribe to or unsubscribe from \\ 
(octet 0) & (1) & (2 - k) \\ \hline 
\end{tabular}
\vspace{2ex}

\begin{tabular}{|p{1.2cm}|p{1.1cm}| p{2cm} | p{1cm}| p{1cm}|}
\multicolumn{5}{c}{ROMANO MQTT PUB Request Message Format} \vspace{1ex} \\ \hline
ROMANO Data Type & MSG Length  & MQTT Topic Length, (m)  & Topic ID & Data to Publish \\ 
(octet 0) & (1) & (2) & (3 - m) & ( m+1 - k) \\ \hline 
\end{tabular}
\vspace{2ex}

\begin{tabular}{|p{1.6cm} | p{2cm} | p{1.5cm}| p{1.5cm}|}
\multicolumn{4}{c}{ROMANO Movement Control Message Format} \vspace{1ex} \\ \hline
ROMANO Data Type & MSG Length & Movement Control Type & Movement Control Data \\ 
(octet 0) & (1) & (2 - 3) & (4 - k) \\ \hline 
\end{tabular}
\vspace{2ex}

\begin{tabular}{|p{1.6cm} | p{2cm} | p{1.5cm}| p{1.5cm}|}
\multicolumn{4}{c}{ROMANO Sensor Data Message Format} \vspace{1ex} \\ \hline
ROMANO Data Type & MSG Length & Sensor Type & Sensor Data \\ 
(octet 0) & (1) & (2 - 3) & (4 - k) \\ \hline
\end{tabular}
\label{table:romano-format}
\end{table}

\begin{table}[!ht]
    \centering
    \caption{Movement Control Types} 
    \begin{tabular}{|p{2cm}|p{2cm}|p{2cm}|}
        \hline
        \multicolumn{2}{|c|}{Movement Control Type} & Movement  \\ \cline{1-2}
        Type            &  Value    &  Control Data \\ \hline \hline
        Move Front      & 0x0000    & Distance \\ \hline
        Move Back       & 0x0001    & Distance \\ \hline
        Move Left       & 0x0002    & Distance \\ \hline
        Move Right      & 0x0003    & Distance \\ \hline
        Rotate Left     & 0x0004    & Angle    \\ \hline
        Rotate Right    & 0x0005    & Angle \\ \hline
    \end{tabular}
   \label{tab:movemnt}

\end{table}
\textbf{Message Formats: }
The communication in ROMANO follows certain message structures described as follows.
The base messaging format in ROMANO is presented in Table~\ref{table:romano_msg}.
The ROMANO data type field dictates the communication type.
The default types are presented in Fig.~\ref{fig:romano} (Right).
The ROMANO MSG Length section denotes the length \emph{(say, k + 1 octets)} of the whole ROMANO message, and the ROMANO Data Section contains the data of variable length. 
Each of the ROMANO message types (except type ROMANO Connection Ack which only uses the Data Type and the Data length section) have their own formats, summarized in Table~\ref{table:romano-format}.
Most of the message formats are self-explanatory except the ROMANO MQTT Publish Request message and Movement Control message.
In MQTT Publish Request messages, the MQTT Topic Length field \emph{(say, m octets)} marks the end of the topic id field (to publish data to) from the beginning of the ROMANO message where the Topic ID section starts at the 3rd octet. The data to publish section in the Publish Request message defines the custom type of data to publish. 
The ROMANO movement control message can be used to control different types of movements. 
We have also defined some basic movement control message type listed in Table~\ref{tab:movemnt}. One can define up to $2^{16}$ different movement control functions (using the allocated 2 octets) with custom arguments. 
In principle, entire sequences of useful movements could be encoded into a single movement control message; for example, a semi-circular motion clockwise around an obstacle, specified by a radius parameter.
ROMANO outputs each movement command to the built-in ROMANO control data mailbox queue (a structure made available for both C and C++) which is serviced by the thread running a robot's movement controller.
The implementation of a movement controller is still independent of ROMANO, but controllers are required to retrieve commands from the ROMANO control data mailbox.





\section{Real Implementation and Experimentation:}

\subsection{Core Implementation:}
\label{sec:impl}
\begin{figure}
    \centering
    \includegraphics[width=0.9\linewidth]{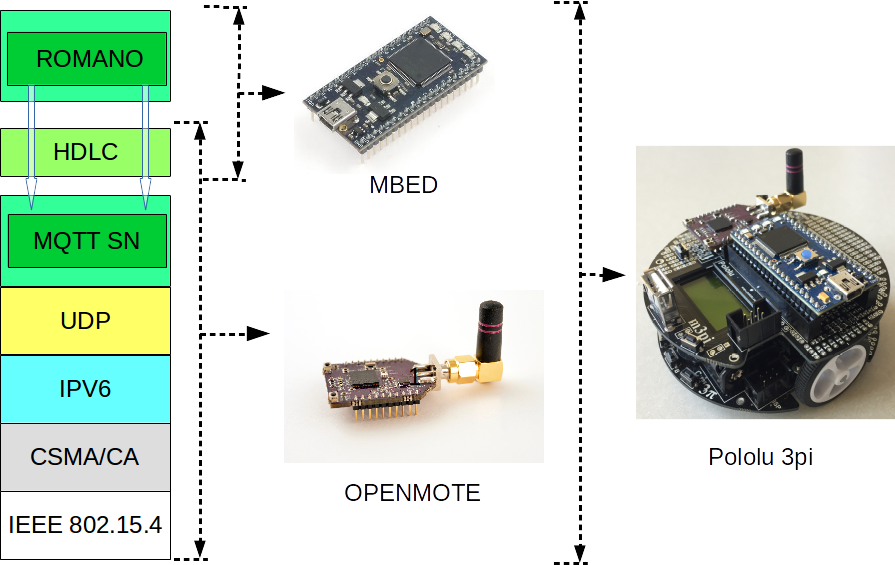}
    \caption{ROMANO Implementation Stack on Pololu 3pi}
    \label{fig:m3pi_romano}
\end{figure}

In this section, we present our real implementation of the proposed ROMANO protocol in a testbed of five cheap, low power, and commercially available robots called Pololu 3pi~\cite{pololu}.
A 3pi, illustrated in Fig.~\ref{fig:m3pi_romano}, also comes with an expansion board that can accommodate an XBee form factor device for IEEE 802.15.4 communication and an mbed board. 
For communication, we use a commercially available product for IoT called the OpenMote~\cite{Openmote}, and for the mbed device, we use the LPC1768~\cite{mbed_ref}.
We choose this particular set of hardware due to the following reasons.
(1) These devices are compatible with each other and have very low computation power, very small communication energy consumption, small form factor, and also comparably low cost.
(2) These devices form the base of our ultimate aim of developing a scalable, portable, cheap, open-source wireless robotic IoT testbed that will be used for research on low power robotics with a major focus on communication and networking.

In this testbed, we implemented the ROMANO in a modular distributed manner over an mbed and an OpenMote. 
The mbed device does not have a radio, but it has more processing power and memory than the OpenMote. The OpenMote comes with a radio but does not have enough GPIO pins and memory to act as the robot controller. 
Thus, we implemented ROMANO across both devices with controllers on the mbed and the communication software stack on the OpenMote with UART bridging data between the two.
For reliable UART communication, we have implemented a low power version of the reliable data transfer protocol High-level Data Link Control (HDLC)~\cite{gelenbe1978performance}.
Our stack implementation is illustrated in Fig.~\ref{fig:m3pi_romano}.
For software development, we use a well-known open source OS for IoT called RIOT-OS~\cite{riot} on the OpenMote and the open-source real-time operating system MBED-OS 5~\cite{mbedos} on the mbed.
Note that, for implementing the ROMANO protocol, one device with both a radio and microcontroller is sufficient. 

\begin{figure}[!ht]
    \centering
    \includegraphics[width=0.85\linewidth]{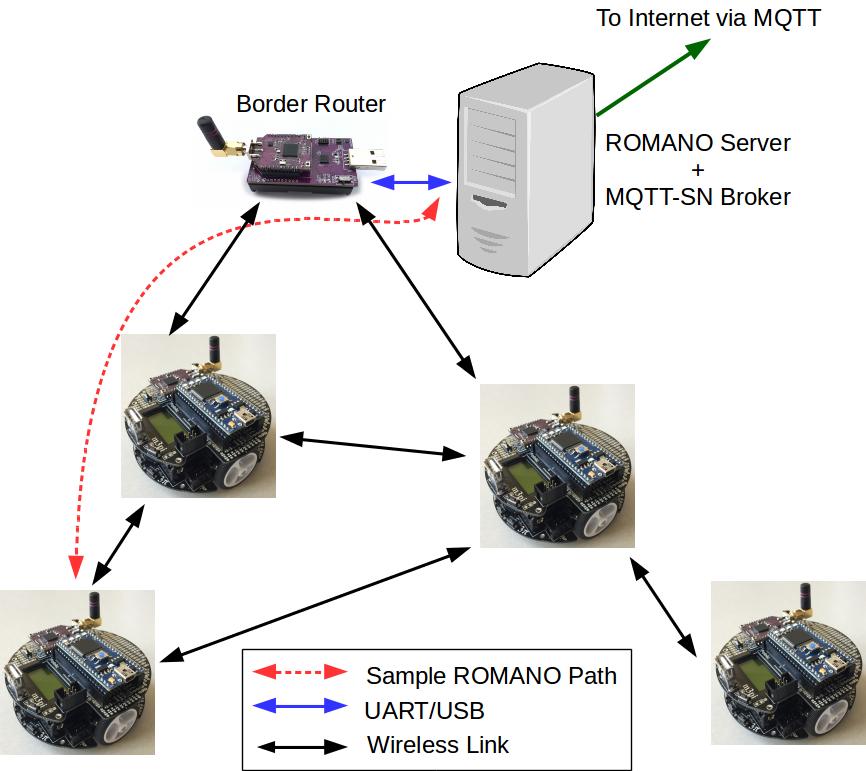}
    \caption{Our Testbed Architecture of ROMANO}
    \label{fig:m3pi_romano1}
\end{figure}

In our current implementation, the MQTT-SN broker is running on a Raspberry Pi running Raspbian with an OpenMote connected via USB to act as the 802.15.4/6LoWPAN border router. All the 3pis are connected to the broker as well as the internet via either a direct link or a multihop link to the border router (illustrated in Fig.~\ref{fig:m3pi_romano1}). For routing in the multihop network we use a well-known routing protocol for 802.15.4 networks called RPL~\cite{RPL}. We use IPv6 for addressing instead of IPv4 due to its wide applicability in IoT systems as well as its compatibility with existing IPv4 systems. We performed the experiments using channel 26 of the IEEE 802.15.4 standard to avoid external interference from WiFi networks.
Our code and designs will be made available at \href{https://github.com/ANRGUSC/ROMANO}{https://github.com/ANRGUSC/ROMANO} upon publication.

\subsection{Performance Analysis}

With the experimental setup detailed in Section~\ref{sec:impl}, we performed a set of experiments to analyze the performance of the proposed ROMANO protocol.
In this section, we focus on three important communication/networking aspects in a robotic network: scalability, end to end delay of communication, and throughput. 
We performed a series of stress tests with the experimental setup to find the performance boundaries of ROMANO.
To test message delivery ratio, we ran a set of experiments where the ROMANO server script publishes messages to the connected robots via ROMANO for message generation rates of 1, 10, 20, 50, 75, 100, 200, 300, 400, and 500 Message(s) Per Second (MPS). For each rate, we ran 10 experiments, each with 5000 messages. We find that the message delivery percentage is $\geq 99.5\%$ for a messaging rate of \emph{200 MPS} or less. For higher message generation rates the testbed system fails after a while due to the radio buffer overflow (summarized in Table~\ref{tab:mps}). After careful investigation, we find that this buffer overflow is due to the radio hardware limitations and not due to the limitations of the ROMANO protocol.
This is further justified by the fact that, until the radio buffer overflows, the message delivery ratio is $\approx 99.5\%$. 

\begin{figure}
    \centering
    \includegraphics[width=0.8\linewidth]{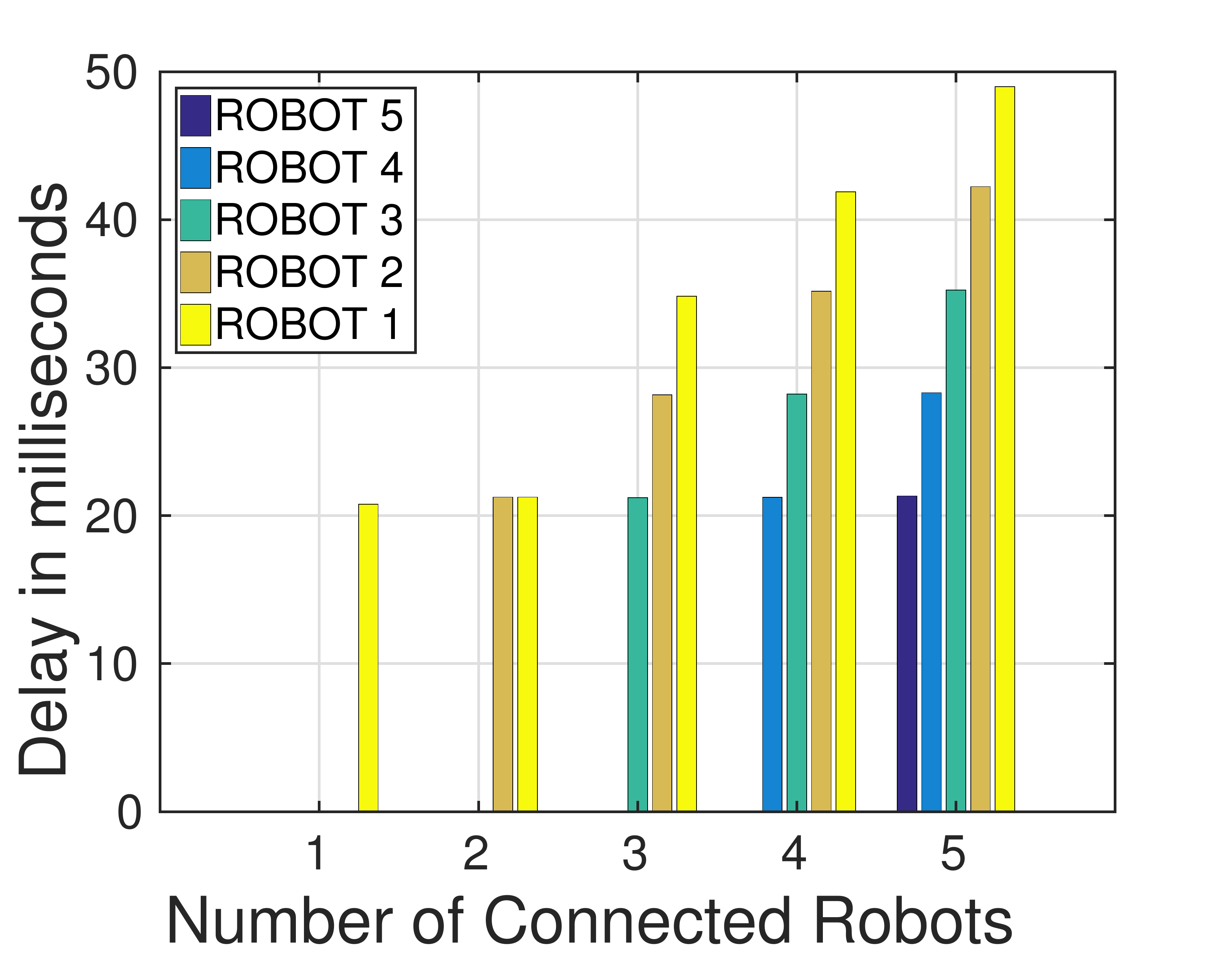}
    \caption{Scalability Analysis of ROMANO Protocol}
    \label{fig:scalab}
\end{figure}

To study the scalabilty as well as the delay, we performed a set of experiments where the server published message at a rate of 20 MPS to the topic ``common'' while we varied the number of robots. 
The results, presented in Fig.~\ref{fig:scalab}, demonstrates that typically the minimum delay is $\approx 20$ms which is justifiable as typical packet transfer time in a 802.15.4 network is $\approx10-20$ms.
Figure~\ref{fig:scalab} also demonstrates that the individual delay experienced by the robots are different. 
This difference in delay is deterministic ($\approx 8$ ms) and due to the operation principle of MQTT-SN (not ROMANO)
where the broker dispatches broadcast messages via sequential unicast messages to one subscriber node at a time.
This suggests that there is a linear relationship between the maximum delay over ROMANO with the number of subscribed robots on that topic. 

In terms of reliability, ROMANO can provide device to broker reliability by using reliability feature of the MQTT-SN with different QoS modes. In the current version, ROMANO does not have any end to end reliability. But one can easily add some level of reliability by using the feature of heartbeat message where each device periodically sends a `heartbeat' message to the `common' topic to inform all nodes about its presence and adding a logic to send messages to a device only if a heartbeat message was received. If a heartbeat message has not been received, the sender can queue it until the destination device rejoins the network.

\begin{table}[!ht]
    \centering
    \begin{tabular}{|c|c|c|p{4cm}|}
    \hline
    \multirow{2}{*}{MPS}     & \multicolumn{2}{|c|}{Message Delivery Ratio} & \multirow{2}{*}{Comments} \\
    \cline{2-3}
            & Minimum       &   Maximum     &          \\ \hline
    100     & $99.9\%$      &   $100\%$     & No Radio Buffer Failure           \\ \hline
    200     & $99.5\%$      &   $100\%$     & No Radio Buffer Failure          \\ \hline
    300     & $44\%$        &   $100\%$     & Radio Buffer Overflow after roughly 2200 message           \\ \hline
    400     & $15\%$        &   $100\%$     & Radio Buffer Overflow after roughly 1300 Messages          \\ \hline
    500     & $13\%$        &   $96\%$      & Radio Buffer Overflow after roughly 600 Messages          \\ \hline

    \end{tabular}
    \caption{Message Delivery Ratio for Different Message Generation Rates}
    \label{tab:mps}
\end{table}
Note that in this paper we do not compare this performance with other protocols as we were unable to find another overlay protocol for robotic networks with a focus on lightweight, low power communications. Of course ROS architecture gives similar features as our proposed ROMANO protocol. Nonetheless, to our knowledge, there exists no generic lightweight low power version of ROS. In this work we do not focus on higher power robots with fully functional computers running Linux with WiFi. The comparison of ROMANO with ROS on such systems is left as a future work. 
Moreover, in the future, we intend to bridge ROMANO to ROS nodes via the MQTT-SN broker in a hierarchical network.

\subsection{Application Implementation Examples}

To test and analyze how our protocol works, we have implemented and tested four different applications detailed as follows. The videos from the experiments will be available at \href{http://tiny.cc/anrg-romano}{http://tiny.cc/anrg-romano}.

\subsubsection{ROMANO for control of a group of robots} 
In this implementation, we mainly use the ROMANO Movement control message format. We have implemented a movement control thread in each robot, which wakes up upon receiving a movement control message via the ROMANO protocol and executes the movement instructions (e.g. move left or right by a fixed amount). The movement control messages can be published to any topics. Therefore, to control a subgroup of robots in a swarm, nodes can either publish to all target ROMANO IDs or publish to a special topic subscribed by only the target subgroup.

\subsubsection{ROMANO Path Copy}
In this application, we implement a very simple system where one 3pi robot runs line-follower code while it shares its telemetry information via ROMANO to a certain topic.
All the other robots listen for the telemetry and use the telemetry to replicate the path.
This illustrates how easily sensor data or any kind of data can be shared between a group of low power low capability robots, and this also shows how ROMANO can be used to control one robot from another.

\subsubsection{ROMANO to Control Peer-to-Peer UDP Communication}
\begin{figure}
    \centering
    \includegraphics[width = 0.8\linewidth]{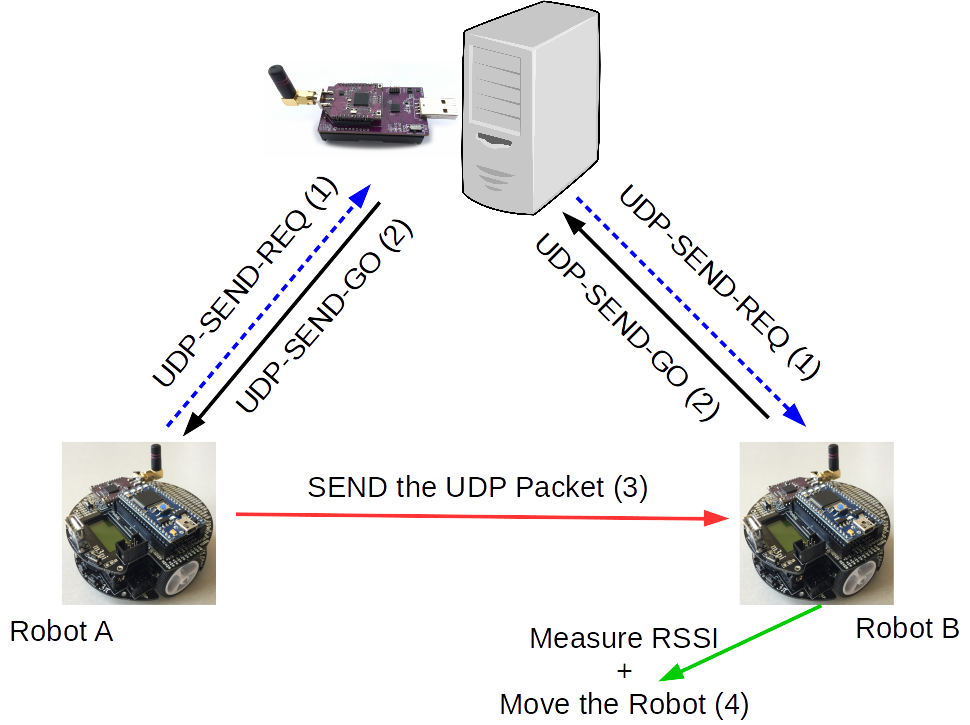}
    \caption{Illustration of using ROMANO to control peer-to-peer UDP communication}
    \label{fig:test3}
\end{figure}

In this implementation, two robots use ROMANO to control the UDP packet transmissions of one another and disperse while the two robots maintain a certain radio communication link quality level.
We add two different custom packet types for this purpose: ``UDP-SEND-REQ'' (0x11) and ``UDP-SEND-GO'' (0x12). To illustrate the application, we describe a sample sequence of events between a robot A and robot B (presented in Fig.~\ref{fig:test3}) as follows. Note the application is implemented on the mbed with the OpenMote sending/receiving messages at the mbed's request. First, robot A publishes a UDP-SEND-REQ message to robot B and receives a UDP-SEND-GO reply. Upon receiving UDP-SEND-GO, the OpenMote on A transmits a broadcast UDP packet. Upon receiving the UDP packet, Robot B's Openmote forwards the Radio Signal Strength Indication (RSSI) of that packet to the mbed using HDLC. If the RSSI is above a user defined threshold $RSSI_{th}$, the mbed moves 3pi B away from A by a fixed step size $d_s$. Similarly, if the value is less than $RSSI_{th}$, the B will move closer to A to improve the link quality. After the movement step, robot A becomes ready to reply to a UDP-SEND-REQ message from robot B (B is sending UDP-SEND-REQ messages at a regular interval until A replies). This process continues indefinitely, and B will follow the same procedure as A. The movement is restricted to forward and backward movement along a black line to leverage the 3pi's reflective sensors for simplicity. The whole process is randomly initiated by one of the robots given the two are within range of their radios.

\subsubsection{ROMANO over Internet}
\begin{figure}
    \centering
    \includegraphics[width = 0.9\linewidth]{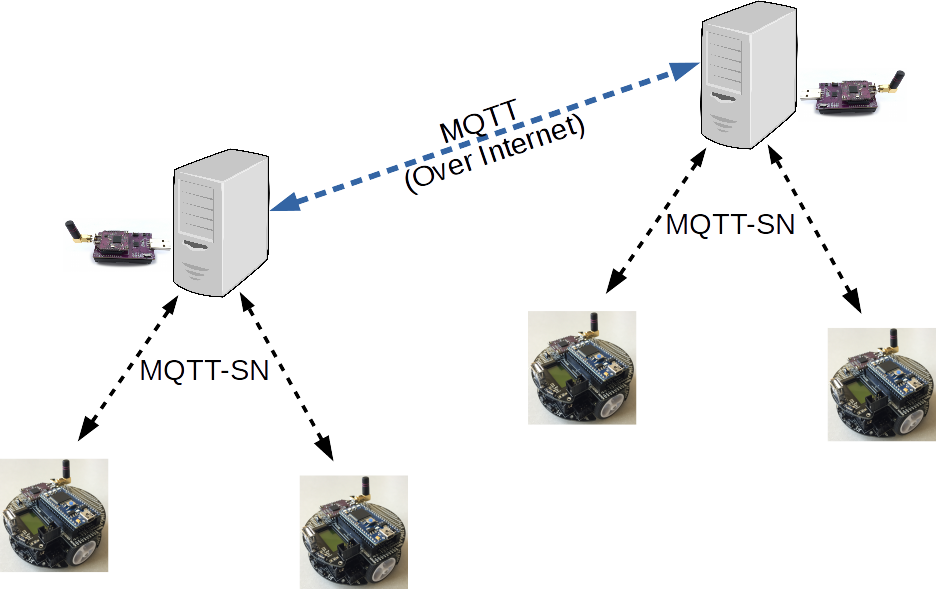}
    \caption{Application of ROMANO for communication between two robotic networks connected over internet.}
    \label{fig:test4}
\end{figure}

Until now, we discussed employing ROMANO over a single 802.15.4 network.
In this application, we show how ROMANO can be used between two networks of robots bridged via the internet.
The 802.15.4 networks use IPv6 address and MQTT-SN protocol for communication while over the internet communication use IPv4 address and MQTT protocol, illustrated in Fig.~\ref{fig:test4}.
It shows that ROMANO is compatible for both IPv6 and IPv4 as well as both MQTT and MQTT-SN.


\section{Conclusion}

In this paper, we presented a proof of concept overlay protocol called ROMANO that works on top of MQTT-SN to provide a light-weight, scalable, and low power communication abstraction for sensing and control in a wireless network of robots. 
We also developed a real system on a robotic testbed consisting of five Pololu 3pi robots and performed a set of evaluations for the proposed ROMANO protocol. 
Through a set of four different application implementations, we demonstrate how the ROMANO protocol can help in the research and development in wireless network protocols for robots.  
However, there are a lot of research questions that remain to be answered.
The current broker in our experimental setup is a centralized one. 
Thus, a major focus of our future work will be on the development of a distributed MQTT-SN broker system where a subset of the robots act as the brokers without the need of a central broker. 
Another potential direction is to formally define control loops over the ROMANO network where one robot of the network controls another arbitrary robot in the network. 
Lastly, a detailed comparative analysis with ROS for robots with high computation power is planned for future work.

\bibliographystyle{unsrt}
\bibliography{ref}

\end{document}